\documentclass[runningheads]{llncs}
\usepackage{makeidx}
\usepackage{graphicx}
\usepackage{amsmath,amssymb} 
\usepackage{color}

\newcommand\myparagraph[1]{\vspace{0pt}\noindent\textbf{#1}\quad}
\usepackage[center]{subfigure}
\usepackage{tabularx}
\usepackage{booktabs}
\usepackage{rotating}
\usepackage{multirow}

\newcommand{\seppl}{\textit{Separate people}}
\newcommand{\sinppl}{\textit{Single fully visible people}}
\newcommand{\dtfull}{\textit{Dense trajectories (DT)}}
\newcommand{\dt}{\textit{DT}}
\newcommand{\gt}{\textit{GT}}
\newcommand{\gtfull}{\textit{GT~single~pose~(GT)}}
\newcommand{\gtt}{\textit{GT-T}}
\newcommand{\gttfull}{\textit{GT~single~pose~+~track~(GT-T)}}

\newcommand{\pstfull}{\textit{PS~single~pose~+~track~(PS-T)}}
\newcommand{\psm}{\textit{PS-M}}
\newcommand{\psmfull}{\textit{PS~multi-pose~(PS-M)}}
\newcommand{\psmdt}{\textit{PS-M~+~DT (features)}}
\newcommand{\psmdtfig}{\textit{PS-M~+~DT}}
\newcommand{\psmfdt}{\textit{PS-M~filter~DT}}
\newcommand{\psmdtclass}{\textit{PS-M~+~DT (classifiers)}}
\newcommand{\pose}{\textit{Pose}}
\newcommand{\occl}{\textit{Occlusion}}
\newcommand{\vp}{\textit{Viewpoint}}
\newcommand{\pl}{\textit{Part~length}}
\newcommand{\trunc}{\textit{Truncation}}
\newcommand{\numdtfull}{\textit{\# dense trajectories (\#~DT)}}
\newcommand{\numdt}{\textit{\# DT}}
\newcommand{\numdtbodyfull}{\textit{\# dense trajectories on body (\#~DT~body)}}
\newcommand{\numdtbody}{\textit{\#~DT~body}}
\newcommand{\msfull}{\textit{Motion speed (MS)}}
\newcommand{\ms}{\textit{MS}}
\newcommand{\msbodyfull}{\textit{Motion speed on body (MS~body)}}
\newcommand{\msbody}{\textit{MS~body}}
\newcommand{\numppl}{\textit{\# people}}

\begin{document}
\pagestyle{headings}
\mainmatter

%
%
\def\DAGM12SubNumber{133}  

%
%
\title{Fine-grained Activity Recognition \\with Holistic and Pose based Features}

%
%
%
\titlerunning{Fine-grained Activity Recognition \\with Holistic and Pose based Features}
\authorrunning{Leonid~Pishchulin, Mykhaylo~Andriluka, and Bernt~Schiele}

\author{
Leonid Pishchulin$^1$ 
\and Mykhaylo Andriluka$^{1,2}$ 
\and Bernt Schiele$^1$ 
}
\institute{
Max Planck Institute for Informatics, Germany
\and
Stanford University, USA\\
}

\maketitle

\begin{abstract}
  Holistic methods based on dense trajectories
  \cite{wang13ijcv,wang13iccv} are currently the de facto standard for
  recognition of human activities in video. Whether holistic
  representations will sustain or will be superseded by higher level
  video encoding in terms of body pose and motion is the subject of an
  ongoing debate \cite{Jhuang:2013:TUA}. In this paper we aim to
  clarify the underlying factors responsible for good performance of
  holistic and pose-based representations.  To that end we build on
  our recent dataset~\cite{andriluka14cvpr} leveraging the existing
  taxonomy of human activities. This dataset includes $24,920$ video
  snippets covering $410$ human activities in total. Our analysis
  reveals that holistic and pose-based methods are highly
  complementary, and their performance varies significantly depending
  on the activity. We find that holistic methods are mostly affected
  by the number and speed of trajectories, whereas pose-based methods
  are mostly influenced by viewpoint of the person. We observe
  striking performance differences across activities: for certain
  activities results with pose-based features are more than twice as
  accurate compared to holistic features, and vice versa.  The best
  performing approach in our comparison is based on the combination of
  holistic and pose-based approaches, which again underlines their
  complementarity.





\end{abstract}

\section{Introduction}

In this paper we consider the task of human activity recognition in
realistic videos such as feature movies or videos from YouTube. We
specifically focus on how to represent activities for the purpose of
recognition. Various representations were proposed in the literature,
ranging from low level encoding using point trajectories
\cite{wang13ijcv,wang13iccv} to higher level representations using
body pose trajectories~\cite{Jhuang:2013:TUA} and collection of action
detectors \cite{Sadanand:2012:ABH}. At the high level human activities
can often be accurately characterized in terms of body pose, motion,
and interaction with scene objects. Representations based on such high
level attributes are appealing as they allow to abstract the
recognition process from nuisances such as camera viewpoint or person
clothing, and facilitate sharing of training data across
activities. However, articulated pose estimation is a challenging and
non-trivial task that is subject of ongoing
research~\cite{pishchulin13cvpr,pishchulin13iccv,yang12pami,dantone13cvpr,sapp13cvpr}. Therefore,
state-of-the-art methods in activity recognition rely on holistic
representations~\cite{LMSR08,duchenne09iccv,wang13ijcv,wang13iccv}
that extract appearance and motion features from the entire video and
leverage discriminative learning techniques to identify information
relevant for the task.

Recent results on the JHMDB dataset \cite{Jhuang:2013:TUA} suggest
that state-of-the-art pose estimation methods might have reached
sufficient accuracy to be effective for activity
recognition. Motivated by these results, we further explore holistic
and pose based representations aiming for much broader scale and
coverage of activity classes. To that end we employ our recent ``MPI
Human Pose'' dataset~\cite{andriluka14cvpr}. Compared to $21$ activity
classes considered in~\cite{Jhuang:2013:TUA} the ``MPI Human Pose''
dataset includes $410$ activities and more than an order of magnitude
more images ($\sim32$K in JHMDB vs. over 1M images in ``MPI Human
Pose''). ``MPI Human Pose'' aims to systematically cover a range of
activities using an existing taxonomy \cite{Ainsworth:2011:CPA}. This
is in contrast to existing datasets \cite{Kuehne11,Soomro:2012:UCF}
that typically include ad-hoc selections of activity classes.
Using the rich labelling of people provided with ``MPI Human Pose'' we
evaluate the robustness of holistic and pose based representations to
factors such as body pose, viewpoint, and body-part occlusion, as well
as to the number and speed of dense trajectories covering the person.

This paper makes the following contributions. First, we perform a
large-scale comparison of holistic and pose based features on the
``MPI Human Pose'' dataset. Our results complement the findings in
\cite{Jhuang:2013:TUA}, indicating that pose based features indeed
outperform holistic features for certain cases. However we also find
that both types of features are complementary and their combination
performs best. Second, we analyze factors responsible for success and
failure, including number and speed of trajectories, occlusion,
viewpoint and pose complexity.

\myparagraph{\textbf{Related work.}} There is a large body of work on
human activity recognition and its review is out of the scope of this
paper. Instead we point to the respective surveys
\cite{VishwakarmaA13,Cardinaux:2011:VBT:2010465.2010468} and
concentrate on the methods most relevant for this work.

Holistic appearance based features combined with the
Bag-of-Words representation
\cite{LMSR08,duchenne09iccv,wang13ijcv,wang13iccv,Jhuang:2013:TUA} are
considered the de facto standard for human activity recognition in
video. Many methods create discriminative feature representations of a
video by first detecting spatiotemporal interest points
\cite{chakraborty11iccv,laptev05ijcv} or sampling them densely
\cite{WUKLS09} and then extracting various feature descriptors in the
space-time volume. Most commonly used feature descriptors are
histograms of oriented gradients (HOG) \cite{Dalal2005CVPR},
histograms of flow (HOF) \cite{dalal06eccv} or Harris 3D interest
points~\cite{laptev05ijcv}. In this paper we examine the recent
holistic approach \cite{wang13ijcv,wang13iccv} which tracks dense
feature points and extracts strong appearance based features
along the trajectories. This method achieves state-of-the-art results
on several
datasets. Other holistic approaches include template
based~\cite{rodriguez08cvpr} or graph based methods constructing
graphs from a spatiotemporal segmentation of the
video~\cite{brendel11iccv}.

Another line of research explores ways of higher level video encoding
in terms of body pose and
motion~\cite{Ferrari:2008:PSS,kumar11iccv,rohrbach12cvpr,Jhuang:2013:TUA}. The
intuition there is that many activities exhibit characteristic body
motions and thus can reliably be described using human body pose based
features. Pose based activity recognition was shown to work
particularly well in images with little clutter and fully visible
people \cite{kumar11iccv}. However, in more challenging scenarios with
frequent occlusions, truncations and complex poses, body features
significantly under-perform holistic appearance based
representations~\cite{rohrbach12cvpr}. Recently, it was shown
that body features extracted from detected joint positions outperform
holistic methods, and their combination did not improve over using
body features only~\cite{Jhuang:2013:TUA}. However, these conclusions
were made on a subset of the HMDB dataset~\cite{Kuehne11}, where
actions with global body motion are performed by isolated and fully
visible individuals -- a setting that seems well suited for pose
estimation methods. In this work we examine a wide range of underlying
factors responsible for good performance of body based and holistic
methods. In contrast to~\cite{Jhuang:2013:TUA} our analysis on
hundreds of activity classes reveals that holistic and pose based
methods are highly complementary, and their performance varies
significantly depending on the activity.


We build our analysis on our recent ``MPI Human Pose'' dataset
collected by leveraging an existing taxonomy of every day human
activities and thus aiming for a fair coverage. This is in contrast to
existing activity recognition datasets
\cite{Kuehne11,Soomro:2012:UCF,MLS09,liu09cvpr,rodriguez08cvpr} that
typically include ad-hoc selections of activity classes.
A large number of activity classes ($410$) and more than an order of
magnitude more images compared to \cite{Jhuang:2013:TUA} facilitate
less biased evaluations and conclusions.



\section{Dataset}
In order to analyze the challenges for fine-grained human activity
recognition, we build on our recent publicly available ``MPI Human
Pose'' dataset \cite{andriluka14cvpr}. The dataset was collected from
YouTube videos using an established two-level hierarchy of over 800
every day human activities. The activities at the first level of the
hierarchy correspond to thematic categories, such as ``Home repair'',
``Occupation'', ``Music playing'', etc., while the activities at the
second level correspond to individual activities, e.g. ``Painting
inside the house'', ``Hairstylist'' and ``Playing woodwind''. In total
the dataset contains $20$ categories and $410$ individual activities
covering a wider variety of activities than other datasets, while its
systematic data collection aims for a fair activity coverage. Overall
the dataset contains $24,920$ video snippets and each snippet is at
least $41$ frames long. Altogether the dataset contains over a 1M
frames. Each video snippet has a key frame containing at least one
person with a sufficient portion of the body visible and annotated
body joints. There are $40,522$ annotated people in total. In
addition, for a subset of key frames richer labels are available,
including full 3D torso and head orientation and occlusion labels for
joints and body parts.

\myparagraph{\textbf{Static pose estimation complexity measures.}}
\label{sec:pose-compl} In addition to the dataset, in
\cite{andriluka14cvpr} a set of quantitative complexity measures
aiming to asses the difficulty of pose estimation in each particular
image was proposed. These measures map body image annotations to a
real value that relates the complexity of each image w.r.t. each
factor. These complexity measures are listed below.
\vspace{-0.5em}
\begin{enumerate}
\item \pose: deviation from the mean pose on the entire dataset.
\item \occl: number of occluded body
parts.
\item \vp: deviation of 3D torso rotation from the
frontal viewpoint.
\item \pl: deviation of body part lengths from the
mean part lengths.
\item \trunc: number of truncated body parts.
\end{enumerate}
\myparagraph{\textbf{Novel motion specific complexity measures.}}
\label{sec:motion-compl}
We augment the above set with the measures assessing
the amount of motion present in the scene.
\vspace{-0.7em}
\begin{enumerate}
\item \numdtfull: total number of DT computed by \cite{wang13iccv}.
\item \numdtbodyfull: number of DT on body mask.
\item \msfull: mean over all trajectory displacements in the video.
\item \msbodyfull: \ms~extracted on body mask.
\item \numppl: number of people in the video.
\end{enumerate}
\section{Methods}
\label{section:methods}
In order to analyze the performance on the challenging task of
fine-grained human activity recognition, we explore two lines of
methods that extract relevant features. The first line of methods
extracts holistic appearance based features and is represented by the
recent ``Dense Trajectories'' method \cite{wang13ijcv} which achieves
state-of-the-art performance on several datasets. The second line of
methods computes features from locations of human body joints
following the intuition that body part configurations and motion
should provide strong cues for activity recognition. We now describe
both types of methods and their combinations.
\subsection{Dense trajectories (DT)} 
DT computes histograms of oriented gradients
(HOG)~\cite{Dalal2005CVPR}, histograms of flow (HOF) \cite{LMSR08},
and motion boundary histograms (MBH) \cite{dalal06eccv} around densely
sampled points that are tracked for 15 frames using median filtering
in a dense optical flow field. In addition, $x$ and $y$ displacements
in a trajectory are used as a fourth feature. We use a publicly
available implementation of the improved DT method \cite{wang13iccv},
where additional estimation removes some of the trajectories
consistent with camera motion. Following \cite{wang13iccv} we extract
all features on our data and generate a codebook for each of the four
features of 4K words using k-means from a million of sampled
features, and stack $L_2-$normalized histograms for learning.

\subsection{Pose-based methods}
It has recently been shown that body features provide a strong signal
for recognition of human activities on a rather limited set of 21
distinctive full body actions in monocular rgb video sequences
\cite{Jhuang:2013:TUA}. We thus investigate the usefulness of body
features for our task where the variability of poses and granularity
of activities is much higher. We explore different ways of obtaining
body joint locations and extract the body features using the code
kindly provided by~\cite{Jhuang:2013:TUA}.
We use the same trajectory length of $7$ frames with a step size of
$3$, generate a codebook of $20$ words for each descriptor type and
finally stack the $L_2-$normalized histograms for learning. We now
present different ways of obtaining body joint locations.

\myparagraph{\textbf{\gtfull.}} We directly use the ground truth
locations of body joints in the key frame to compute single pose based
features. As some of the body parts may be truncated, we compute
features only for the present parts.

\myparagraph{\textbf{\gttfull.}}  As the ground truth information is
not available for the rest of the frames in a sequence, we approximate
the positions of body joints in the neighboring frames by tracking the
joints using sift-based tracker~\cite{rohrbach12cvpr}. 
The tracker is initialized with correct positions of body joints, and
thus provides reliable tracks of joints in the local temporal
neighborhood.

\myparagraph{\textbf{\pstfull.}} It is not realistic to expect the
ground truth information to be available at test time in real world
scenarios. We thus replace the given body joint locations by
automatically estimated ones using the publicly available
implementation \cite{yang12pami}. This efficient method is based on
pictorial structures (PS) and obtained good performance on the ``MPI
Human Pose'' \cite{andriluka14cvpr}.

\myparagraph{\textbf{\psmfull.}} Using the method \cite{yang12pami}
also allows to obtain joint locations independently in each frame of a
sequence without using the sift tracker. Notably, the same method was
shown by \cite{Jhuang:2013:TUA} to outperform the holistic approach.
\subsection{Combinations of holistic and body based methods}

As the holistic \dt~approach does not extract any pose information,
and pose based methods do not compute any appearance features, both
are potentially complementary. Thus we expect that an activity
recognition system will profit from their combinations. We investigate
two ways of combining the methods.

\myparagraph{\textbf{\psmdt.}} We perform a \textit{feature} level
fusion of both \dt~and \psm~by matching both types of features
independently to the respective codebooks and then stacking the
normalized histograms into a single representation.

\myparagraph{\textbf{\psmdtclass.}} We also investigate a
\textit{classifier} level fusion. To do so we first run
pre-trained~\dt~and \psm~classifiers~(see Sec.~\ref{sec:results})
independently on each sequence and stack the scores together into a
single feature vector.

\myparagraph{\textbf{\psmfdt.}} Another way of combining both types of
methods is using estimated joint locations to filter the trajectories
computed by \dt. We first estimate poses in all video frames and
generate a binary mask using the union of rectangles around detected
body parts for all single top detections per frame. We then only
preserve the trajectories overlapping with the mask in all frames.

\section{Analysis of activity recognition performance}
\label{sec:results}
In this section we analyze the performance of holistic and pose based
methods and their combinations on the challenging task of fine-grained
human activity recognition with hundreds of activity classes.

\renewcommand{\subfigtopskip}{0pt}%
\renewcommand{\subfigcapskip}{5pt}%
\renewcommand{\subfigbottomskip}{15pt}%
\renewcommand{\subfigcapmargin}{0pt}%
\begin{figure*}[t]
  \small 
  
  \centering \bgroup \tabcolsep 1.5pt
  
  \renewcommand{\arraystretch}{0.2} \subfigure[\# examples/activity on
  \seppl]{\includegraphics[trim=1.4cm 6cm 2.2cm 6cm,
    clip=true,width=.31\linewidth]{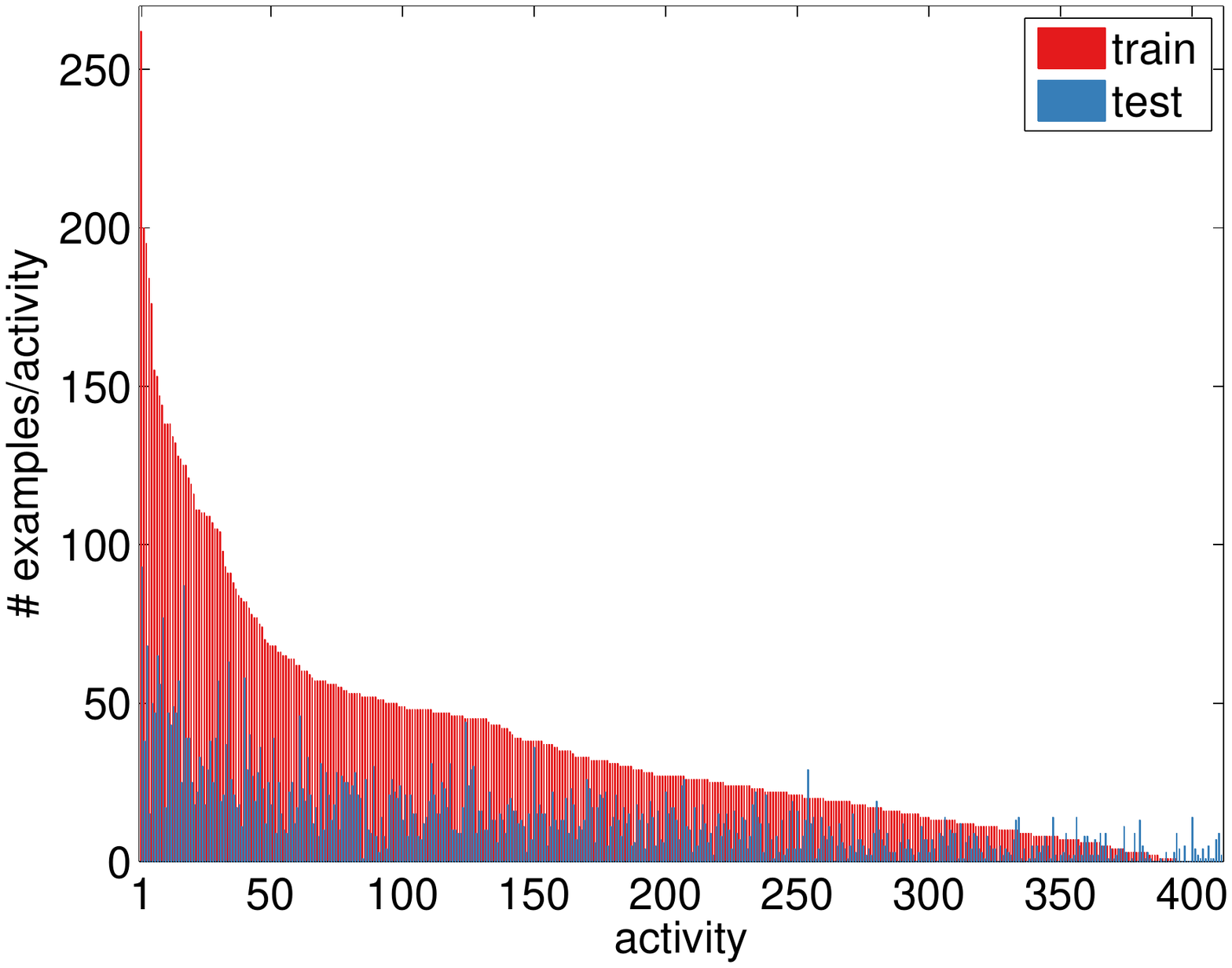}
    \label{fig:act-num-ex}
  }
  \subfigure[Performance (mAP) on \seppl]{\includegraphics[trim=1.4cm 6cm 2.2cm 6cm,
    clip=true,width=.31\linewidth]{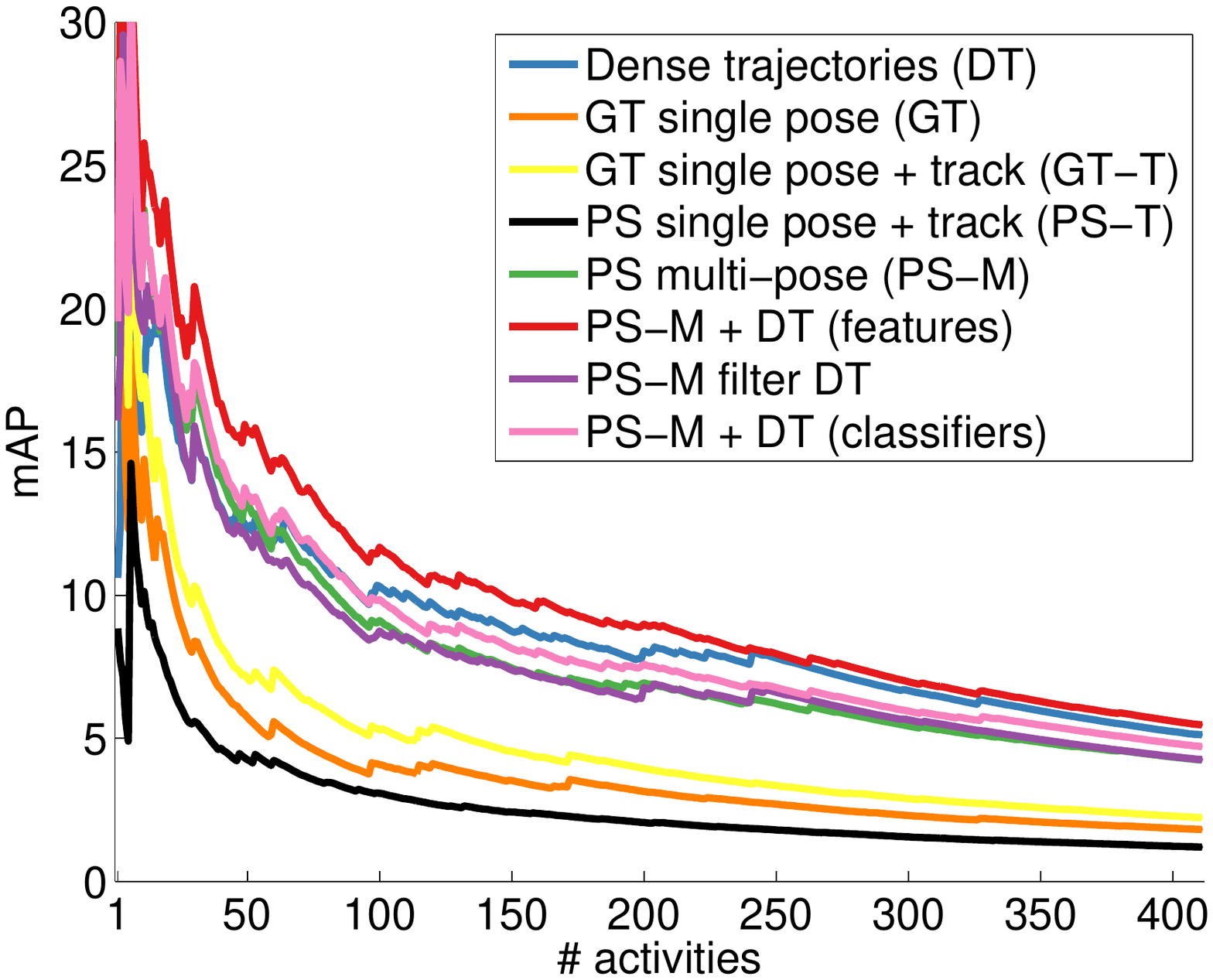}
    \label{fig:total-resuls:separate-ppl}
  }
  \subfigure[Performance (mAP) on \sinppl]{\includegraphics[trim=1.4cm 6cm 2.2cm 6cm,
    clip=true,width=.31\linewidth]{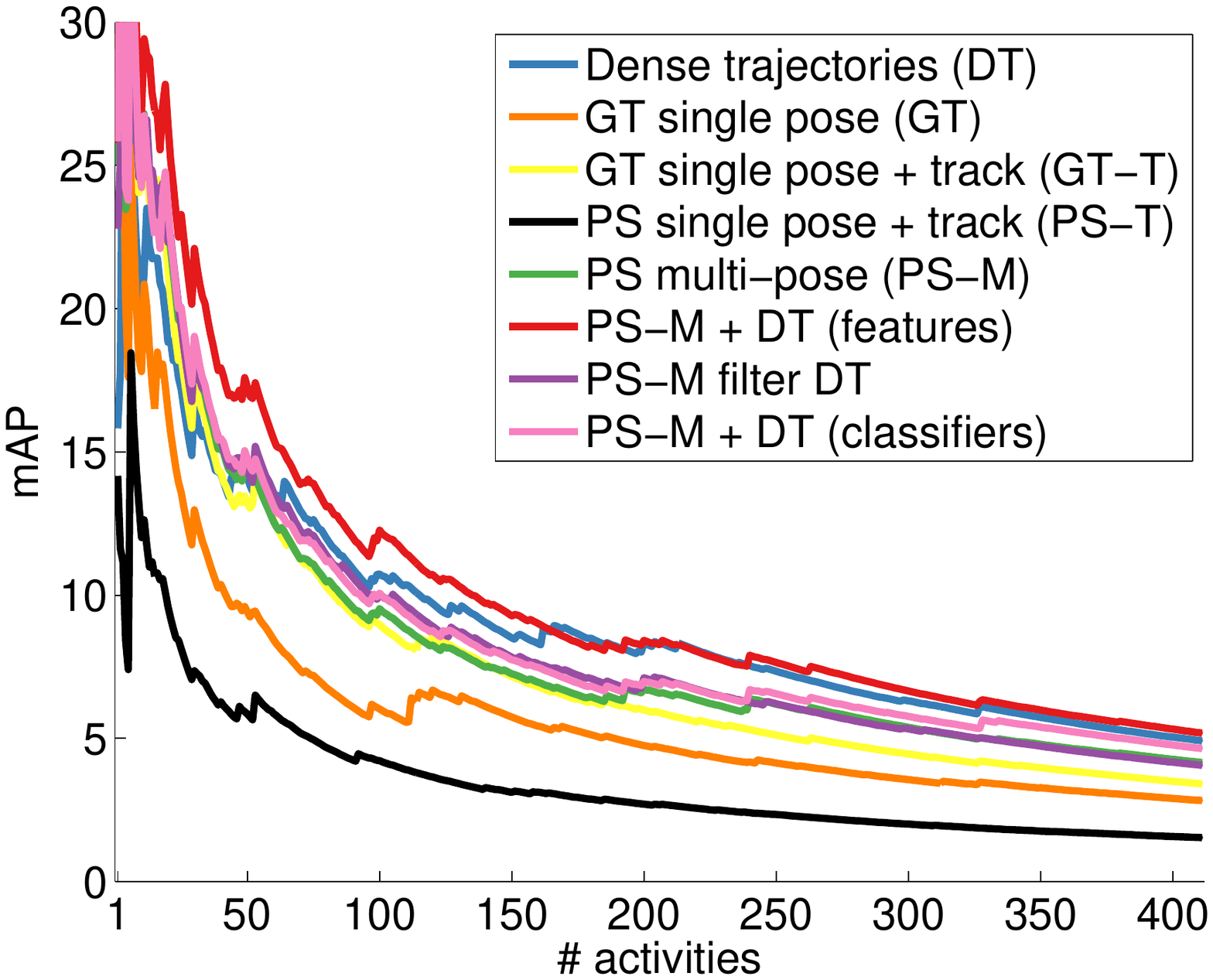}
    \label{fig:total-resuls:single-ppl}
  }
  \vspace{-2em}
  \caption{Dataset statistics and performance (mAP) as a function
    training set size. Shown are (a) number of training/testing
    examples/activity in \seppl~subset; performance on (b) \seppl~and
    (c) \sinppl. Best viewed in color.}
  \label{fig:total-resuls}
  \egroup
  \vspace{-1.3em}
\end{figure*}

\myparagraph{\textbf{Data splits.}} As main test bed for our analysis,
a split with videos containing sufficiently separated individuals
\cite{andriluka14cvpr} is used. This restriction is necessarily for
using the pose estimation method \cite{yang12pami}. This \seppl~split
contains $411$ activities with $15244$ training and $5699$ testing
video snippets. Fig.~\ref{fig:act-num-ex} shows statistics of the
training and testing videos per activity. Notably, the videos may
still contain multiple people and some body parts may be truncated by
a frame border. To rule out the confusion potentially caused by the
presence of multiple truncated people, we define a subset of the
test set from \seppl. This subset contains $2622$ videos with
exactly one fully visible person per video. This \sinppl~setup is
inspired by \cite{Jhuang:2013:TUA} and is favorable for the pose
estimation method \cite{yang12pami} designed to predict body joints of
fully visible people.

\myparagraph{\textbf{Training and evaluation.}} We train activity
classifiers using feature representations described in
Sec.~\ref{section:methods} and ground truth activity labels. In
particular, we train one-vs-all SVMs using mean stochastic gradient
descent (SGD) \cite{rohrbach11cvpr} with a $\chi^2$ kernel
approximation \cite{DBLP:conf/cvpr/VedaldiZ10}. At test time we
perform one-vs-all prediction per each class independently and report
the results using mean Average Precision
(AP)~\cite{pascal-voc-2007}. When evaluating on a subset, we always
report the results on the top $N$ activity classes arranged w.r.t
training set sizes.

\subsection{Overall performance}
We start the evaluation by analyzing the performance on all activity
classes.

\myparagraph{\textbf{\seppl.}} It can be seen from
Fig.~\ref{fig:total-resuls:separate-ppl} that performance is
reasonable for a relatively small number of classes (the typical case
for many activity recognition datasets), but quickly degrades for a
large number of classes, clearly leaving room for improvement of
activity recognition methods.

We observe that \dtfull~alone outperforms all pose based methods
achieving 5.1\% mAP. Expectedly, \gtfull~performs worst (1.8\%
mAP). Although \gt~uses ground truth joint positions to extract body
features, they are computed in a single key frame, thus ignoring
motion. Expectedly, adding motion via sift tracking (\gttfull)
improves the results to 2.2\% mAP. Replacing ground truth by predicted
joint locations (\pstfull) results in a performance drop (1.2\% vs
2.2\% mAP) due to unreliable initialization of the tracker by
imperfect pose estimation. Surprisingly, \psmfull~significantly
improves the results, achieving 4.2\% mAP. It shows that performing
body joint predictions in each individual frame is more reliable than
simple tracking. Interestingly, the feature level fusion
\psmdt~noticeably improves over \dt~alone and classifier level fusion
\psmdtclass, achieving 5.5\% mAP. This shows that both holistic
\dt~and pose based~\psm~methods are complementary. We analyze whether
the complementarity of \dt~comes from the holistic features extracted
on the person or elsewhere in the scene. By restricting the extraction
to the body mask (\psmfdt), we observe a drop of performance
w.r.t. \dt~(4.3\% mAP vs. 5.1\% mAP). It shows that the features
extracted outside of the body mask do contain additional information
which helps to better discriminate between activities in a
fine-grained recognition setting. This intuition is additionally
supported by the similar performance of \psmfdt~w.r.t. \psm. Overall,
we conclude that holistic and pose based methods are complementary and
should be used in a combination for better activity recognition.
 
\myparagraph{\textbf{\sinppl.}} We now analyze the results in
Fig.~\ref{fig:total-resuls:single-ppl}. Although the absolute
performances are higher, which is explained by an easier setting, the
ranking is similar to Fig.~\ref{fig:total-resuls:separate-ppl}. Two
differences are: 1) \gtt~achieves similar performance to
\psm~on many activity classes, but looses in total (3.4\% mAP
vs. 4.2\%~mAP); and 2) \psmfdt~is better than both \dt~and \psm~on a
small set of classes, probably because the trajectory features on the
background mostly contribute to confusion on this set of activities.

\myparagraph{\textbf{Differences to \cite{Jhuang:2013:TUA}.}} Our
analysis in a fine-grained activity recognition setting on hundreds of
classes leads to conclusions which go beyond the results
of~\cite{Jhuang:2013:TUA} obtained from much smaller number of classes
from HMDB dataset \cite{Kuehne11}. First, we compare the performance
of \dt~to a larger number of pose based methods and show the superior
performance of~\dt, when evaluated on hundreds of activities. This is
in contrast to \cite{Jhuang:2013:TUA} showing that the pose based
\psm~is better. Second, we discover that holistic \dt~and pose
based \psm~are complementary and their combination outperforms each of
the approaches alone. This contradicts the conclusions of
\cite{Jhuang:2013:TUA} which does not show any improvement when
combining \dt~and \psm. Finally, we showed that using the trajectories
restricted to body only degrades the performance, which suggests that
the context adds to the discrimination between activity classes.

\subsection{Analysis of activity recognition challenges}
\begin{figure*}[t]
\small
\centering
\bgroup
\tabcolsep 1.5pt
\renewcommand{\arraystretch}{0.2}

\subfigure[\dt]{
\includegraphics[trim=1.5cm 6cm 2.0cm 7.3cm,
clip=true,width=.31\linewidth]{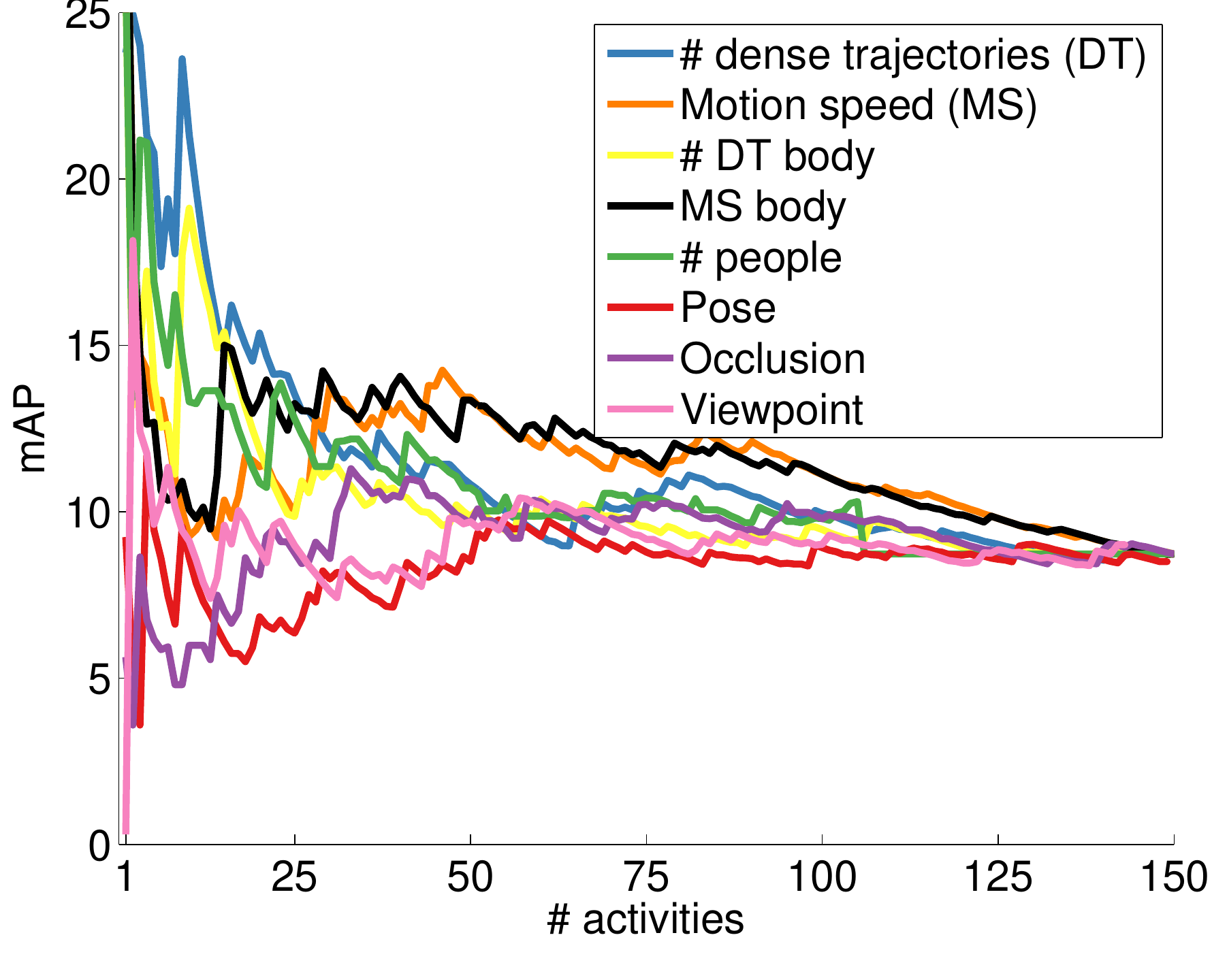}
\label{fig:act-compl-dt}
}
\subfigure[\psmfull]{
\includegraphics[trim=1.5cm 6cm 2.0cm 7.3cm,
clip=true,width=.31\linewidth]{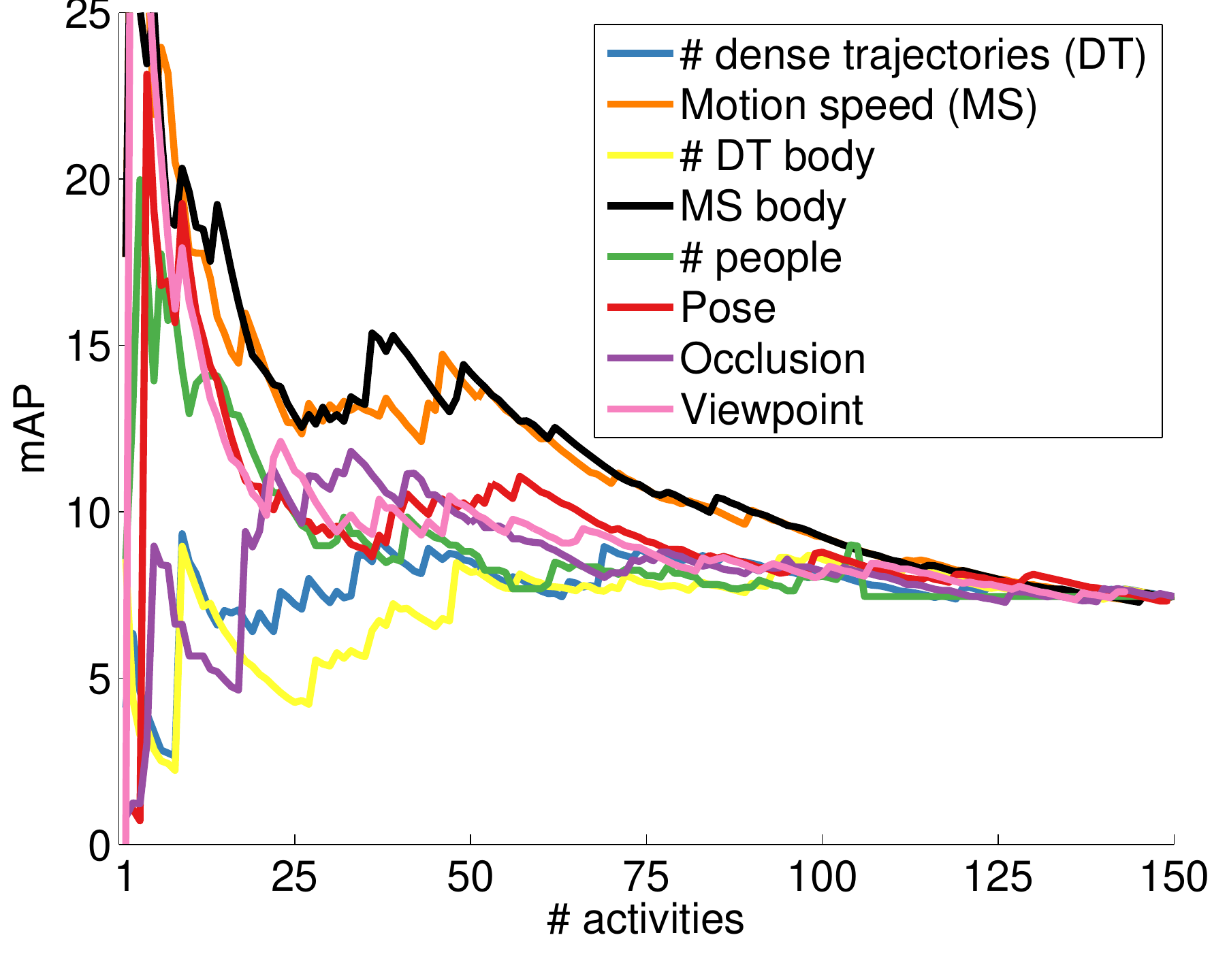}
\label{fig:act-compl-psm}
}
\subfigure[\psmdt]{
\includegraphics[trim=1.5cm 6cm 2.0cm 7.3cm,
clip=true,width=.31\linewidth]{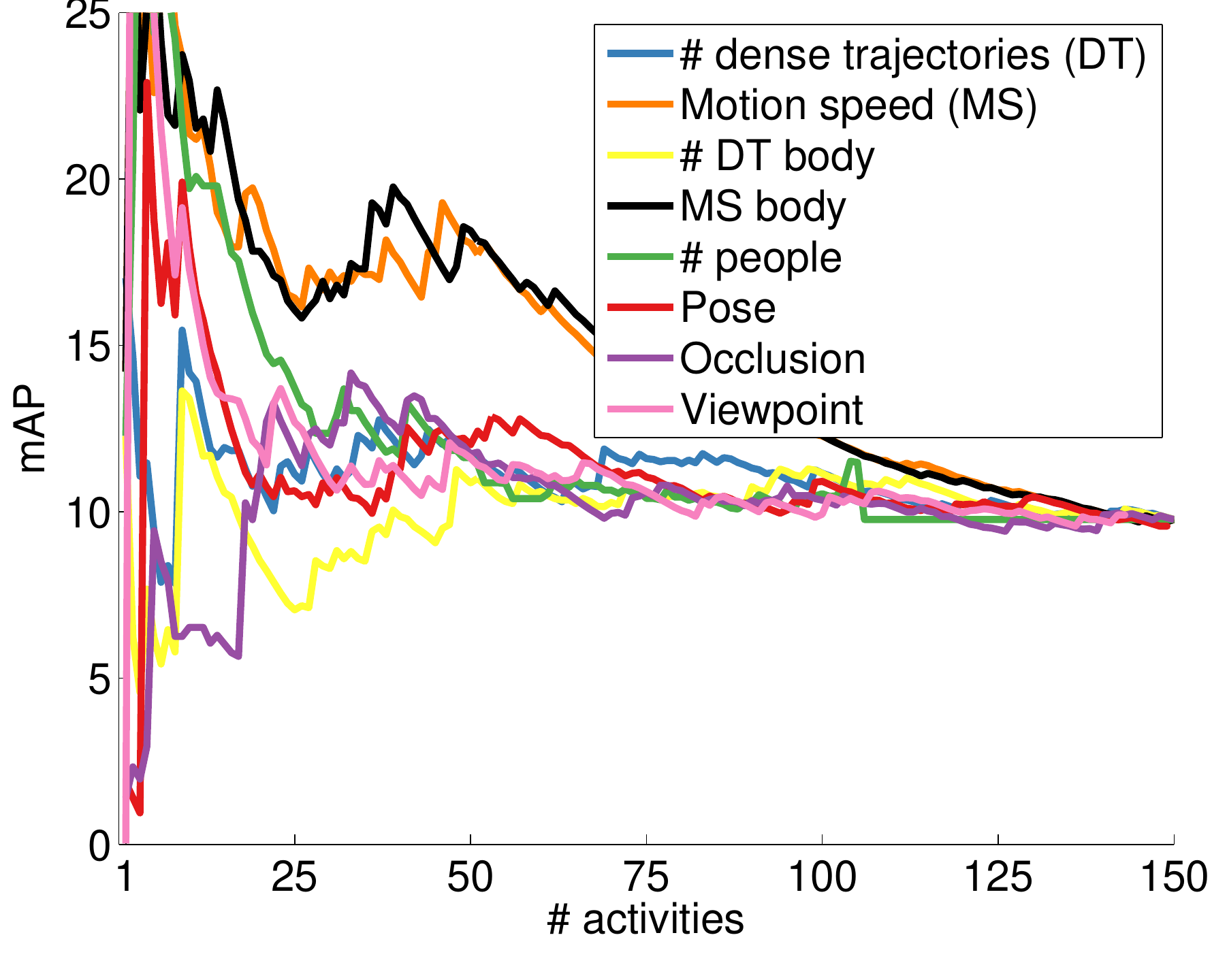}
\label{fig:act-compl-psmdt}
}

\vspace{-2em}
\caption{Performance (mAP) on a subset of 150 activities from
  \seppl~as a function of the complexity measures. Best viewed in
  color and with additional zooming.}
\label{fig:act-compl}
\egroup
\vspace{-1.3em}
\end{figure*}

After analyzing the overall recognition performance on all classes, we
explore which factors affect the performance of best performing
holistic \dt, pose based \psm~and combination \psmdt~of both
methods. We use the complexity measures $1-3$ specific for static pose
estimation and our novel $1-5$ motion specific complexity measures
described in Sec.~\ref{sec:motion-compl}. To make the evaluation
consistent with the rest of the experiments, we compute the average
complexities for the whole activity class and use the obtained real
values to rank the classes. This is in contrast
to~\cite{andriluka14cvpr} which computes the measures per single pose
and thus operates on individual instance level. To visualize the
performance, we sort the activities using the pose related complexity
measures in \textit{increased} complexity order, and motion related
complexity measures in the \textit{decreased} order. As performance
may still be dominated by the training set size when only few examples
are available, we restrict the evaluation to the 150 largest activity
classes. This corresponds to a slice at 150 in
Fig.~\ref{fig:total-resuls:separate-ppl}. The results are shown in
Fig.~\ref{fig:act-compl}. \renewcommand{\subfigtopskip}{0pt}%
\renewcommand{\subfigcapskip}{5pt}%
\renewcommand{\subfigbottomskip}{15pt}%
\renewcommand{\subfigcapmargin}{0pt}%

\myparagraph{\textbf{\dtfull.}} Analyzing the results in
Fig.~\ref{fig:act-compl-dt} we observe that a high number of dense
trajectories everywhere in the video (\numdt) and on human body
(\numdtbody) leads to the best performance of the \dt~method. Also, we
notice that high motion speed (\ms, \msbody) is an indicative factor
for good recognition results. Surprisingly, \dt~performs better on
activities with a high number of people (\numppl). This is explained
by the fact that more people potentially produce more motion, which is
a positive factor for \dt. On the other hand, being close to the
average pose (\pose) and having little occlusion (\occl) hurts
performance. The former is not very surprising, as the average pose is
common to many activities, which makes it more difficult for \dt~to
capture distinctive features. We discover that activities with little
occlusion often contain either little motion (e.g. ``sitting, talking
in person'') or fine-grained motion (e.g. ``wash dishes'') and thus
are hard for \dt.

\myparagraph{\textbf{\psmfull.}} We now analyze
Fig.~\ref{fig:act-compl-psm} and observe several distinctive
differences w.r.t. which factors mostly affect the performance of
\psm. It can be seen that \ms~and \msbody~have stronger effect on
\psm~compared to \dt, and the higher the speed, the better the
performance. In order to better understand this nontrivial trend, we
analyze which activities happen to produce highest \msbody. We note
that those are sports, dancing and running related activities, for
which the pose estimation performance of \cite{yang12pami} is above
average (cf. Fig.~7 in \cite{andriluka14cvpr}). Also, these activities
exhibit characteristic body part motions and can successfully be
encoded using body features. At the other end of the \msbody~scale are
the activities with low fine-grained motion, related to home repair,
self care and occupation, for which the pose estimation performance is
much worse. \pose~and \vp~strongly affect the performance as well, as
frontal upright people whose pose is close to the mean pose are easier
for pose estimation. This is again in contrast to \dt, where the
performance is not noticeably affected by \vp~and even drops in case
of low \pose. Surprisingly, high \numppl~positively affects
\psm. Looking at top ranked activities, we notice that many of them
are related to active group exercises or team sports, such as
``aerobic'' and ``frisbee'', or to simple standing postures, such as
``standing, talking in person''. Body features can again be
successfully used to encode the corresponding motions. On the other
hand, we observe that the high \numdt~and \numdtbody~hurts
performance, which is in contrast to the \dt~method. We observe that
for high \numdtbody~many activities correspond to water related
activities, such as ``fishing in stream'', ``swimming, general'',
``canoeing, kayaking''. Interestingly, the presence of water leads to
high \numdt~and characteristic motions captured by \dt. At the same
time \psm~fails due to unreliable pose estimation caused by complex
appearance and occlusions.

\myparagraph{\textbf{\psmdt.}} The differences for \dt~and
\psm~methods suggest that both methods are complementary. We analyze
in Fig.~\ref{fig:act-compl-psmdt} which factors affect the performance
of \psmdt. It can be seen that positively affecting factors are either
positive for both \dt~and \psm~(\ms, \msbody, \numppl), or positive
for \psm~only (\pose, \vp). In contrast to \psm~the high
\numdt~slightly improves the performance, while high \numdtbody~does
not hurt as much. Expectedly, \vp~hurts performance as it does for
both \dt~and \psm. This shows the complementarity of both \dt~and \psm
and leaves room for improvement in finding better ways of combining
both methods.

\subsection{Detailed analysis on a subset of activities}

After analyzing the factors affecting the results by different
methods, we conduct a detailed analysis on a smaller set of the top
$15$ activities from \seppl.  

\tabcolsep 1.0pt
\begin{table}[tbp]
  \scriptsize
  \centering
  \begin{tabular}{@{}lccccccccc@{}}
    \toprule
    & yoga, &bicycl.,& skiing, & cooking & skate-   & rope    &
    softball, & forestry \\
    & power &mount.   & downh. & or food& board. & skip.& general         &\\
    \midrule
    Dense trajectories (DT) & 10.6 & 14.5 & 51.9 & 0.5 & 11.4 & 36.0 & \textbf{12.7} & 8.4 \\
    
    GT single pose (GT) & 22.3 & 26.5 & 7.5 & 1.8 & 3.4 & 51.2 & 2.2 & 1.4 \\
    
    GT single pose + track (GT-T) & \textbf{37.0} & 28.0 & 10.9 & \textbf{2.6} & 4.6 & 69.2 & 3.6 & 1.2 \\
    
    PS single pose + track (PS-T) & 8.8 & 6.6 & 6.0 & 1.3 & 1.7 & 63.1 & 1.6 & 1.8 \\
    
    PS multi-pose (PS-M) & 18.3 & 34.0 & 27.3 & 2.6 & 17.2 & \textbf{90.5} & 3.0 & 5.2 \\
    
    PS-M + DT (features) & 19.6 & \textbf{40.7} & 32.9 & 2.2 & \textbf{19.5} & 88.7 & 3.9 & 7.2 \\
    
    PS-M filter DT & 16.1 & 20.4 & \textbf{52.2} & 0.8 & 13.5 & 55.7 & 4.2 & \textbf{10.6} \\
    
    \toprule
    \toprule
    
    & carpentry, & bicycl., & golf & rock    & ballet, & aerobic & resist. & total \\
    & general    & racing     &      & climb.& modern  & step    &
    train.   & \\
    \midrule
    Dense trajectories (DT) & 5.5 & 5.5 & 33.0 & \textbf{41.5} & 12.7 & 24.5 & \textbf{16.5} & 19.0 \\
    
    GT single pose (GT) & 2.7 & 7.1 & 36.1 & 2.3 & 1.0 & 1.1 & 1.4 & 11.2 \\
    
    GT single pose + track (GT-T) & 2.8 & 8.7 & 25.3 & 8.9 & 1.7 & 3.3 & 1.3 & 13.9 \\
    
    PS single pose + track (PS-T) & 5.3 & 0.5 & 14.7 & 1.2 & 2.8 & 11.1 & 1.6 & 8.5 \\
    
    PS multi-pose (PS-M) & 3.4 & 8.6 & 47.9 & 4.7 & 22.9 & 10.4 & 7.2 & 20.2 \\
    
    PS-M + DT (features) & 5.0 & 12.1 & \textbf{51.9} & 14.4 & \textbf{23.7} & 17.1 & 14.4 & \textbf{23.5} \\
    
    PS-M filter DT & \textbf{6.1} & \textbf{15.5} & 15.9 & 38.6 & 7.1 & \textbf{25.8} & 9.6 & 19.5 \\
    \bottomrule
    \end{tabular}
  \vspace{0.1em}
  \caption[]{Activity recognition results (mAP) on 15 largest
    classes from \seppl.}
  \label{tab:detail-res:separate-ppl}
  \vspace{-3.5em}
\end{table}

\begin{figure*}[t]
\small
\centering
\bgroup
\tabcolsep 1.5pt
\renewcommand{\arraystretch}{0.2}
{
\scriptsize
\begin{tabular}{@{}c ccc ccc @{}}
  \toprule
  & cooking or & canoeing, & carpentry, & sanding & ballet, & aerobic \\
  & food prep. & kayaking  & general    & floors& modern  & step    \\
    \midrule
&&&&&&\\

\begin{sideways}\dt  \end{sideways}&
\includegraphics[width=.145\linewidth]{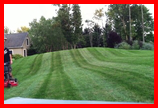}&
\includegraphics[width=.145\linewidth]{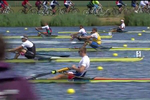}&
\includegraphics[width=.145\linewidth]{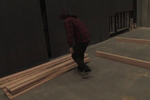}&
\includegraphics[width=.145\linewidth]{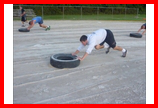}&
\includegraphics[width=.145\linewidth]{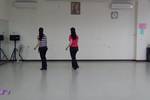}&
\includegraphics[width=.145\linewidth]{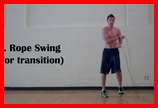}\\
    & mowing lawn, & canoeing, & carpentry, & army type & ballet, & rope \\
    & walking      & kayaking  & general    & training  & modern  & skipping    \\
&&&&&&\\
&&&&&&\\
&&&&&&\\
&&&&&&\\
\begin{sideways}\psm  \end{sideways}&
\includegraphics[width=.145\linewidth]{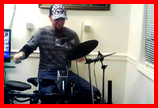}&
\includegraphics[width=.145\linewidth]{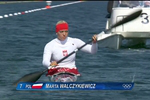}&
\includegraphics[width=.145\linewidth]{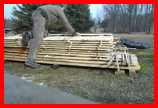}&
\includegraphics[width=.145\linewidth]{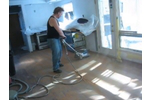}&
\includegraphics[width=.145\linewidth]{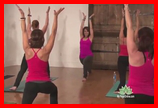}&
\includegraphics[width=.145\linewidth]{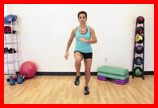}\\
  & playing drums, & canoeing, & carrying, loading, & sanding   & yoga, & circuit \\
  & sitting & kayaking & or stacking wood    & floors& power  & training \\

\begin{sideways}\psmdtfig  \end{sideways}
&
\includegraphics[width=.145\linewidth]{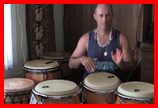}&
\includegraphics[width=.145\linewidth]{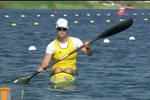}&
\includegraphics[width=.145\linewidth]{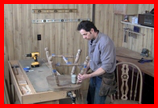}&
\includegraphics[width=.145\linewidth]{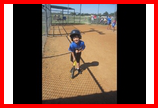}&
\includegraphics[width=.145\linewidth]{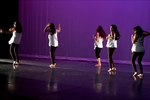}&
\includegraphics[width=.145\linewidth]{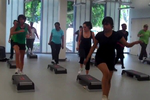}\\
    & drumming & canoeing, & carpentry, & childrens    & ballet, & aerobic \\
    & bongo    & kayaking  & furniture     & games& modern  & step    \\
\end{tabular}
}

\caption{Successful and failure cases on several activity
  classes. Shown are the most confident prediction per class. False
  positives are highlighted in red.}
\label{fig:top-detect}
\vspace{-1.5em}
\egroup
\end{figure*}

 The
results are shown in Tab.~\ref{tab:detail-res:separate-ppl}. None of
the methods outperforms all others on all activities and different
approaches are better on different activities. On average methods
perform well on activities with simple poses and motions e.g. ``rope
skipping'', ``skiing, downhill'' and ``golf'' - typical cases in most
of the current activity recognition benchmarks. However, the
performance of all methods is low for activities with more variability
in motion and poses, e.g. ``cooking'', ``carpentry, general'' and
``forestry''. This leaves room for improvement of current
methods. Analyzing the performance on individual activities, we
observe that for ``yoga, power'' activity \gt~outperforms holistic
\dt~and~\psmfdt~methods (22.3\% vs. 10.6\% and 16.1\% mAP,
respectively) and is better than the pose based \psm~(22.3\%
vs. 18.3\% mAP). It is interesting, as \gt~does not use any motion and
relies on static body features only. The explanation is that the
``yoga, power'' activity contains distinctive body poses and thus can
be reliably captured by \gt, while \psm~fails due to unreliable pose
estimations. It can be seen that in many cases the combination
\psmdt~noticeably outperforms both \psm~and \dt~alone. The differences
are most pronounced for ``bicycling, mountain'', ``bicycling,
racing'', ``skateboarding'' exhibiting characteristic motions, and
``golf'' activity having distinctive body motion and poses. Overall
\psmdt~achieves the best performance of 23.5\% mAP. We visualize
several successful and failure cases of the methods in
Fig.~\ref{fig:top-detect}.

\section{Conclusion}
In this work we address the challenging task of fine-grained human
activity recognition on a recent comprehensive dataset with hundreds
of activity classes. We study holistic and pose based representations
and analyze the factors responsible for their performance. We reveal
that holistic and pose based methods are complementary, and their
performance varies significantly depending on the activity. We found
that both methods are strongly affected by the speed of
trajectories. While the holistic method is also strongly influenced by
the number of trajectories, pose based methods are strongly affected
by human pose and viewpoint. We observe striking performance
differences across activities and experimentally show that the
combination of both methods performs best.

\myparagraph{Acknowledgements.} The authors would like to thank Marcus
Rohrbach and Sikandar Amin for helpful discussions. This work has been
supported by the Max Planck Center for Visual Computing \&
Communication.

\bibliographystyle{splncs03}
\bibliography{biblio}

\end{document}